%% file: main.tex
\documentclass{IEEEtran4PSCC}
\usepackage{amsfonts}

\usepackage[ruled,linesnumbered,noresetcount,vlined]{algorithm2e}
\usepackage{booktabs}

\usepackage{cite}

\ifCLASSINFOpdf
   \usepackage[pdftex]{graphicx}
\else
   \usepackage[dvips]{graphicx}
\fi
\usepackage{xcolor}

\usepackage[cmex10]{amsmath}
\usepackage{url}

\makeatletter
\let\old@ps@headings\ps@headings
\let\old@ps@IEEEtitlepagestyle\ps@IEEEtitlepagestyle
\def\psccfooter#1{%
    \def\ps@headings{%
        \old@ps@headings%
        \def\@oddfoot{\strut\hfill#1\hfill\strut}%
        \def\@evenfoot{\strut\hfill#1\hfill\strut}%
    }%
    \def\ps@IEEEtitlepagestyle{%
        \old@ps@IEEEtitlepagestyle%
        \def\@oddfoot{\strut\hfill#1\hfill\strut}%
        \def\@evenfoot{\strut\hfill#1\hfill\strut}%
    }%
    \ps@headings%
}
\makeatother

\begin{document}
%
\title{Data-Driven Time Series Reconstruction for Modern Power Systems Research}

\author{
\IEEEauthorblockN{Minas Chatzos, Mathieu Tanneau, Pascal Van Hentenryck}
\IEEEauthorblockA{Georgia Institute of Technology\\
\{minas, mathieu.tanneau\}@gatech.edu, pascal.vanhentenryck@isye.gatech.edu}
}


\maketitle

\begin{abstract}
A critical aspect of power systems research is the availability of suitable data, access to which is limited by privacy concerns and the sensitive nature of energy infrastructure.
This lack of data, in turn, hinders the development of modern research avenues such as machine learning approaches or stochastic formulations.
To overcome this challenge, this paper proposes a systematic, data-driven framework for reconstructing high-fidelity time series, using publicly-available grid snapshots and historical data published by transmission system operators.
The proposed approach, from geo-spatial data and generation capacity reconstruction, to time series disaggregation, is applied to the French transmission grid.
Thereby, synthetic but highly realistic time series data, spanning multiple years with a 5-minute granularity, is generated at the individual component level.
\end{abstract}


\input{tex/intro}

\input{tex/reconstruction}

\input{tex/bids}

\input{tex/disaggregation}

\input{tex/extensions}



\bibliographystyle{IEEEtran}
\bibliography{IEEEabrv,refs.bib}
%

\end{document}

%% file: tex/intro.tex
\section{Introduction}
\label{sec:introduction}

A critical aspect of power systems research is the availability of suitable data to, e.g., replicate the operations of a Transmission System Operator (TSO) over multiple days, evaluate stochastic and risk-aware formulations, and/or train machine-learning models.
Indeed, all these applications require a combination of (i) topology information, (ii) time series data at the component level, e.g., historical load at every bus, and (iii) the ability to generate forecasts and quantify uncertainty at each time point in such time series.
Nevertheless, access to real data at this granularity is hindered by the sensitive nature of the energy infrastructure and economic parameters.

Optimal Power Flow (OPF) benchmark sets \cite{Zimmerman2011_MATPOWER,coffrin2014nesta,babaeinejadsarookolaee2019power}, as well as synthetic cases \cite{Birchfield2017_SyntheticNetworks,TAMUData}, provide snapshots of artificial power grids of varying size. These do not include any temporal data, except for a few test cases in \cite{TAMUData} that provide synthetic load time series.
Furthermore, TSOs typically publish, at various time granularity, regional and system-wide load, generation by fuel type, and market-related data \cite{eCO2mix,MISOData,NYISOData,PJMDataMiner}, but no network information. To the best of our knowledge, only the IEEE RTS \cite{Barrows2020_RTS-GMLC} includes both network and spatio-temporally consistent load and renewable output time series data, albeit for a small system and without variability in production costs.

\begin{figure}[!t]
    \centering
    \includegraphics[width=0.8\columnwidth]{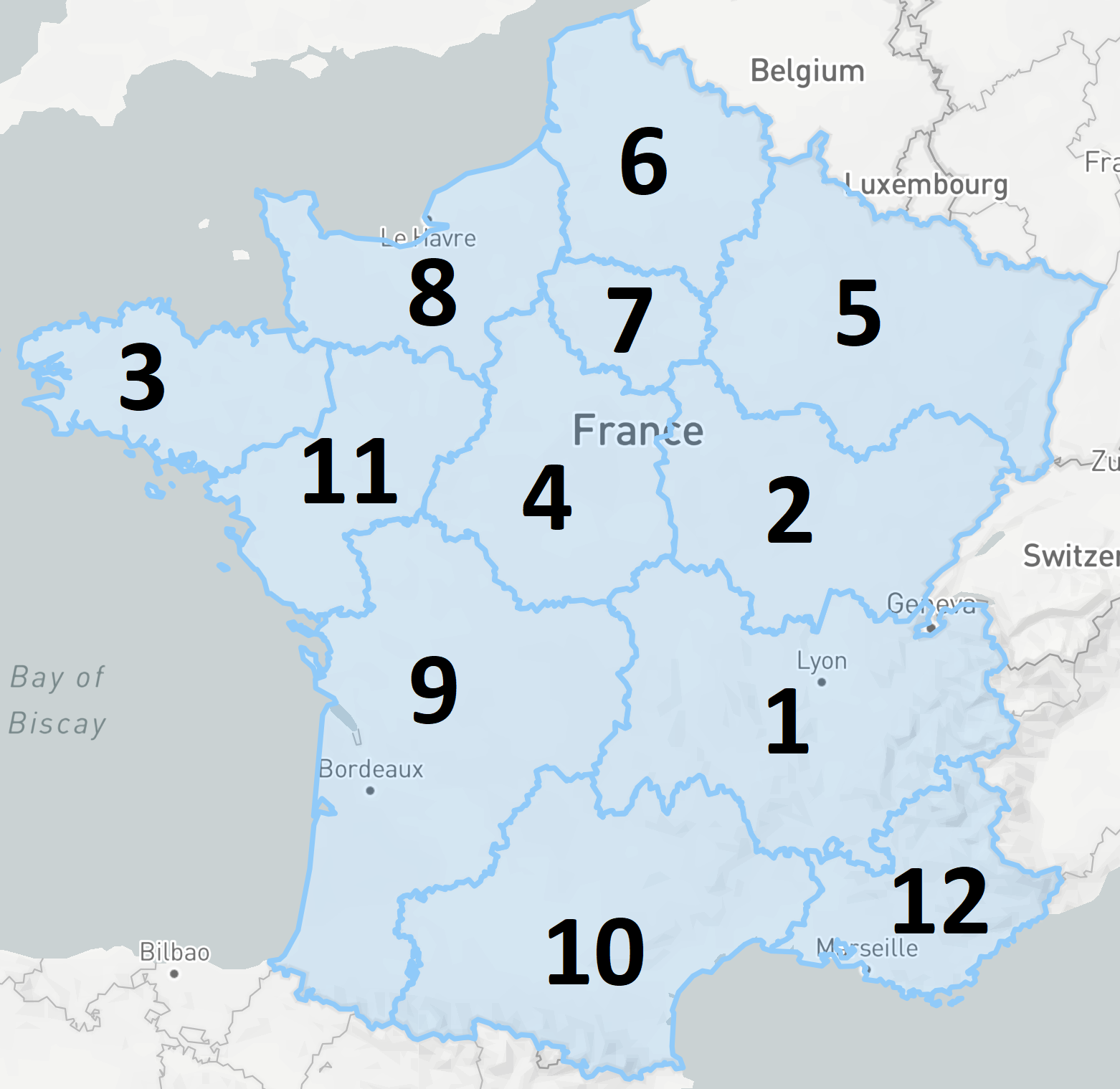}\\
    \caption{The Administrative Regions in Mainland France; see Table \ref{tab:france_regions:indices} for Index Correspondence.}
    \label{fig:france_regions:geojson}
\end{figure}

\begin{figure}[!t]
    \centering
    \includegraphics[width=0.9\columnwidth]{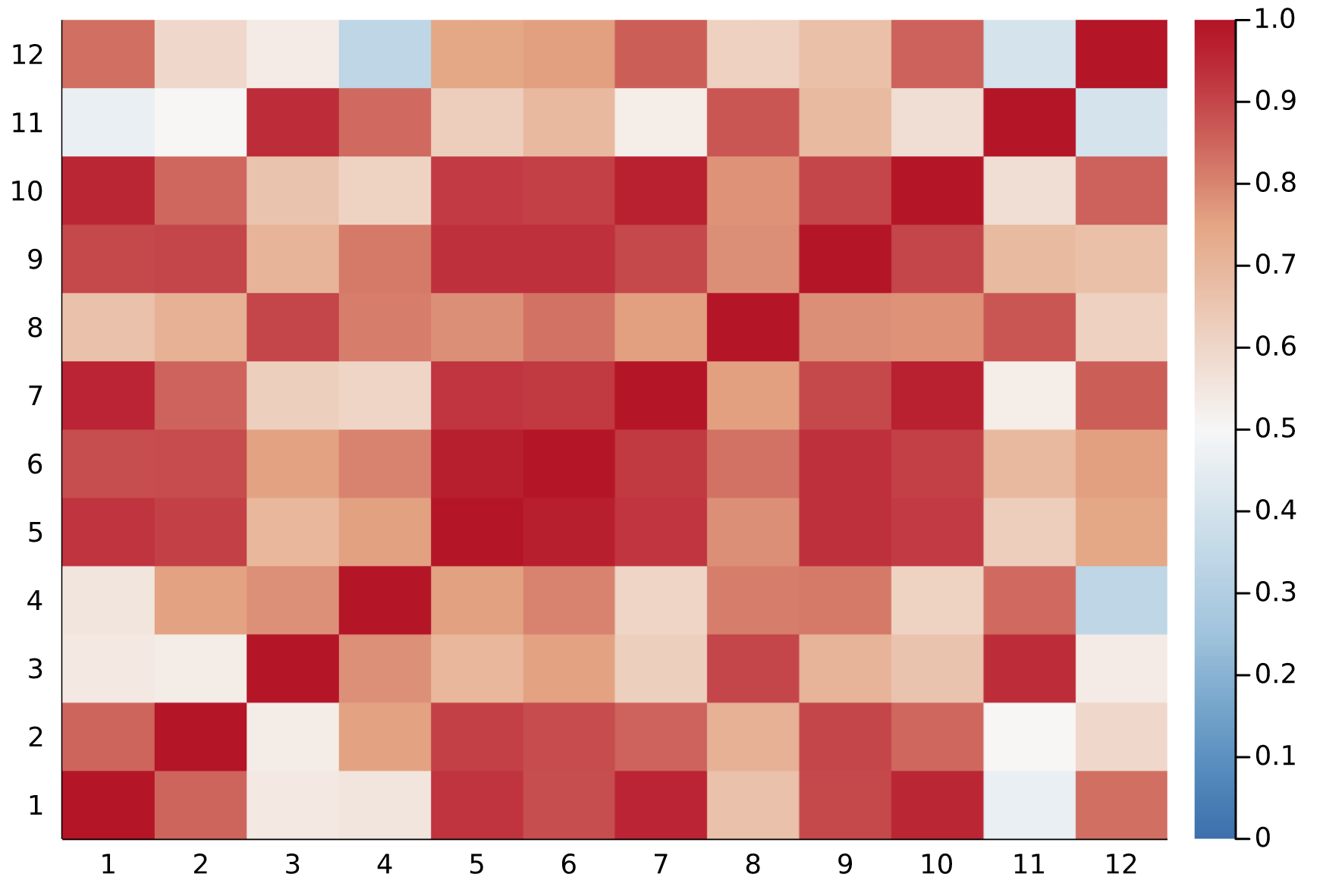}
    \caption{Pearson Correlation Coefficients for Regional Loads on Jan 19, 2018.}
    \label{fig:pearson_correlation:load}
\end{figure}

\begin{figure}[!t]
    \centering
    \includegraphics[width=0.9\columnwidth]{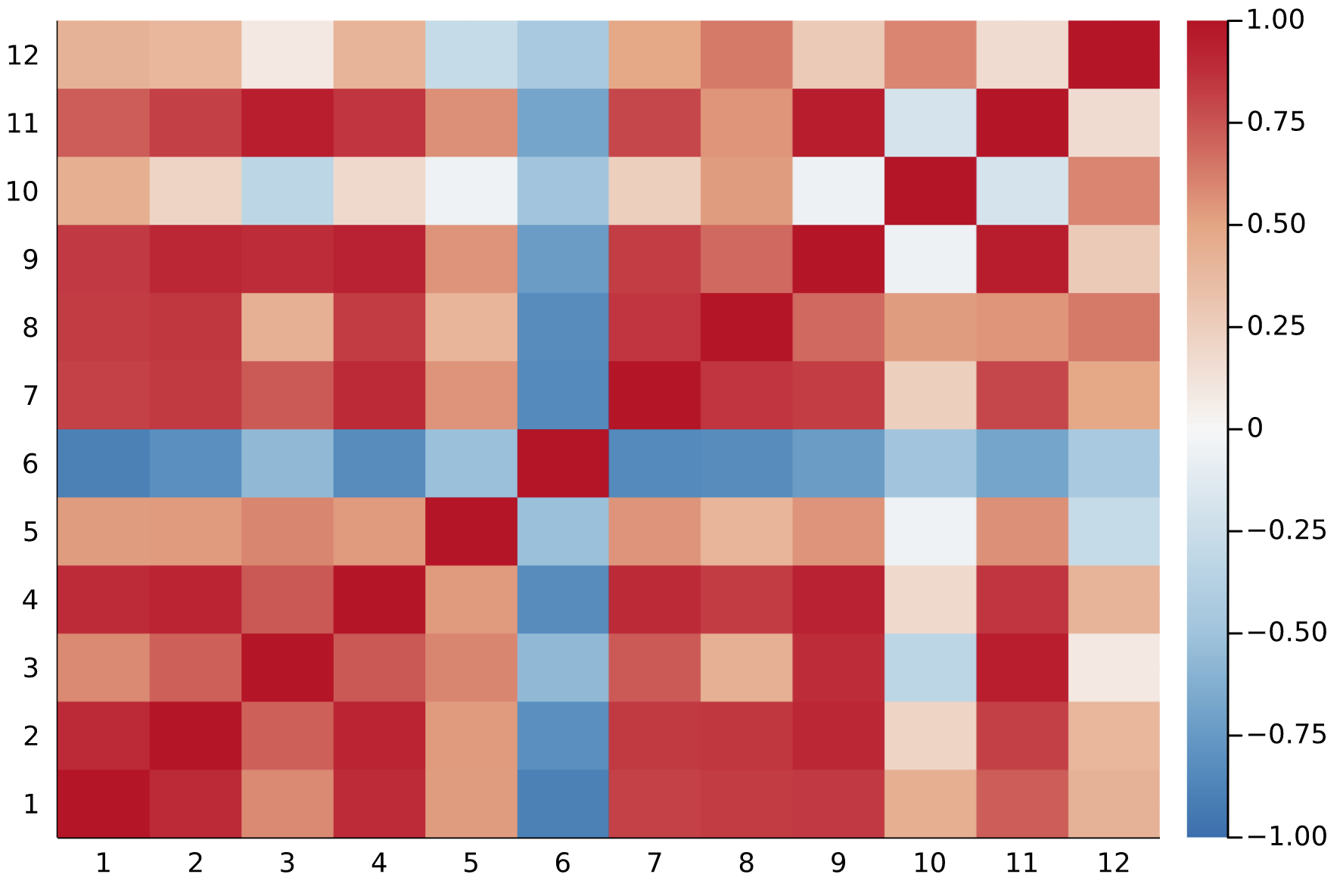}
    \caption{Pearson Correlation Coefficients for Regional Wind Production on Jan 19, 2018.}
    \label{fig:pearson_correlation:wind}
\end{figure}

Thus, to generate training/testing data and/or to sample stochastic scenarios, a growing number of papers \cite{donno2019,Sun2019_SSDiP,Woo2020_RealTimeOPF,Fioretto2020_PredictingACOPF,velloso2020combining,venzke2020learning,pan2020_DeepOPF,Zamzam2020_LearningOptimalACOPFSolutions,Xavier2021_LearningToSolveSCUC,Pineda2021_LearningUCLowHangingFruit} rely on artificially-generated data, which limits the scope of these studies due to various simplifying assumptions regarding the distributions of the synthetic data. In particular, load and renewable production in real-life systems are known to exhibit spatio-temporal correlations \cite{Li2018_LoadModelingSynthetic,Werho2021_ScenarioGenerationWind} due to local weather patterns. To understand these considerations, consider the French transmission grid, whose twelve administrative regions in France are depicted in Figure \ref{fig:france_regions:geojson}, with the index correspondence provided in Table \ref{tab:france_regions:indices}.
Figure
\ref{fig:pearson_correlation:load} (resp. Figure  \ref{fig:pearson_correlation:wind}) shows the inter-regional Pearson correlation coefficients for regional loads (resp. wind productions) on January 19, 2018; this historical data was obtained from \cite{eCO2mix} for every region at 30 minutes granularity.
Unsurprisingly, all regional loads tend to be correlated to one another, with the correlation coefficients higher for neighboring regions, e.g., between Grand-Est (5), Hauts-de-France (6), and Ile de France (7).
Negatively-correlated wind productions may be observed across distant regions. For instance, the wind production in the southern region of Provence Alpes Cote d'Azur (12) is negatively correlated with those of Grand-Est (5), and Hauts-de-France (6), which are both in the Northern part of the country.

\begin{table}[!t]
    \centering
    \caption{Index correspondence for France administrative regions.}
    \label{tab:france_regions:indices}
    \begin{tabular}{clcl}
        \toprule
        Index & Region & Index & Region\\
        \midrule
        1 & Auvergne Rhone-Alpes &
        2 & Bourgogne Franche-Comté \\
        3 & Bretagne &
        4 & Centre-Val de Loire \\
        5 & Grand-Est &
        6 & Hauts-de-France \\
        7 & Ile de France &
        8 & Normandie \\
        9 & Nouvelle Aquitaine &
        10 & Occitanie  \\
        11 & Pays de la Loire  &
        12 & Provence-Alpes-Côte d'Azur \\
        \bottomrule
    \end{tabular}
\end{table}

In \cite{venzke2020learning,pan2020_DeepOPF}, individual loads are sampled independently according to a uniform distribution around the original snapshot; neither renewable production nor cost variability are considered. This sampling strategy does not reflect the spatial correlations between loads in real systems \cite{Li2018_LoadModelingSynthetic}.
Moreover, for large systems, when loads are sampled independently, one can easily verify that the distribution of the system's total load will be concentrated around its expected value; in contrast, the total load of transmission systems may vary by more than $30\%$ within a given day \cite{PJMDataMiner}.

\begin{figure}[!t]
    \centering
    \includegraphics[width=0.95\columnwidth]{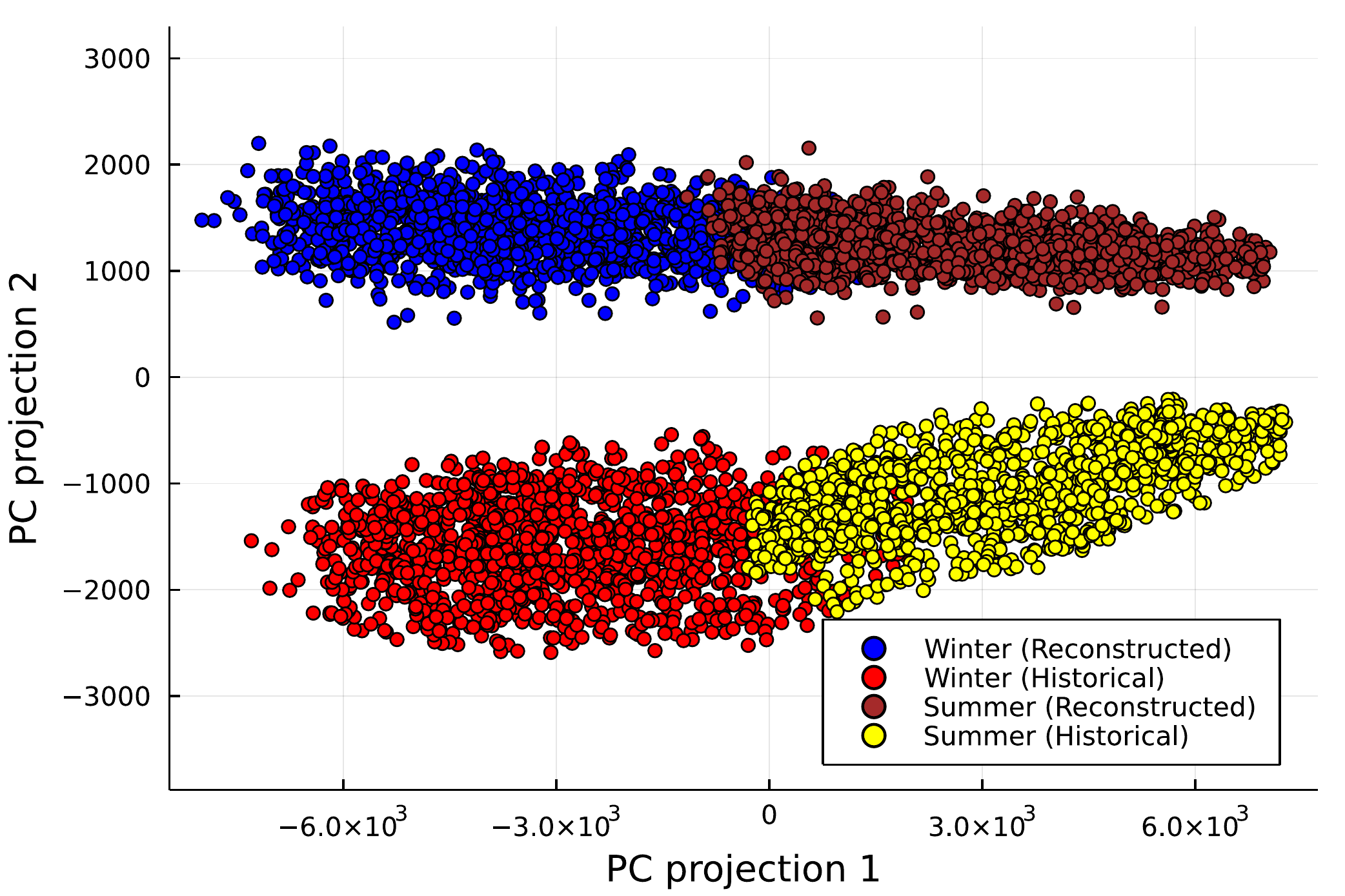}
    \caption{Projected Distribution of Regional Load Profiles Obtained from Historical Data, and from a Uniform Disaggregation as in \cite{donno2019}, using scaling ratios  estimated from a network snapshot and $5\%$ noise.
    The two principal components used for the projection capture $95.6\%$ of the total variance.}
    \label{fig:PCA:nocorr:snap:0.05}
\end{figure}

\begin{figure}[!t]
    \centering
    \includegraphics[width=0.95\columnwidth]{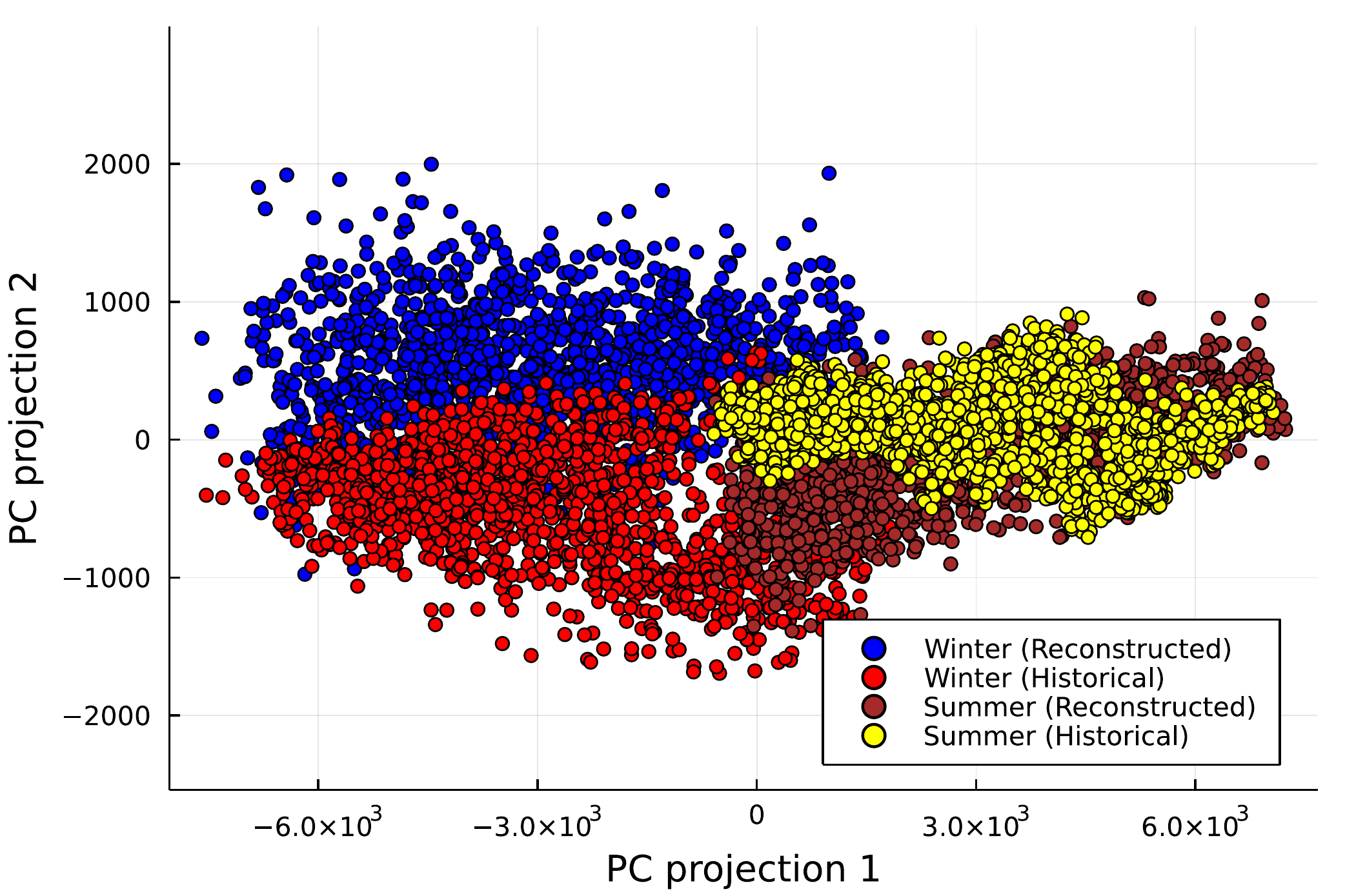}
    \caption{Projected Distribution of Regional Load Profiles Obtained from Historical Data, and from a Uniform Disaggregation as in \cite{donno2019}, using scaling ratios estimated from the previous year's data and $5\%$ noise.}
    \label{fig:PCA:nocorr:hist:0.05}
\end{figure}

In references \cite{donno2019,Sun2019_SSDiP,Woo2020_RealTimeOPF,Fioretto2020_PredictingACOPF,velloso2020combining,Xavier2021_LearningToSolveSCUC,Pineda2021_LearningUCLowHangingFruit}, the system's total load is sampled first, then scaled down to obtain individual load values.
This scaling follows the ratios observed in the reference snapshot, with the addition of a small level of noise.
While this accounts for variations in total load, it does not capture spatial correlations.
Except in \cite{Sun2019_SSDiP}, wherein the authors consider net load, no renewable generation is considered, and only \cite{Xavier2021_LearningToSolveSCUC} takes into account variability in production costs.
According to \cite{donno2019}, \emph{``if a sufficiently high [noise] is set, the sampling strategy we adopted includes the real distribution"}; the validity of this assumption is assessed on the French system.
First, regional and national load profiles are collected from \cite{eCO2mix}, for every 30 minute across January and July 2018.
The historical distribution of regional loads is then compared to that obtained by disaggregating the national load into regional profiles, using a fixed ratio and a multiplicative noise of mean 1 and standard deviation 0.05 (as was used in \cite{donno2019}).
The scaling ratios are estimated from a network snapshot provided by the French TSO, RTE.
To visualize the distributions, this dataset is projected onto its first two principal components, which explain $95.6\%$ of the total variance.
The projected data is shown in Figure \ref{fig:PCA:nocorr:snap:0.05}: each dot is the projection of a 12-dimensional vector containing the twelve regional loads (either historical or reconstructed) at a given time. Figure \ref{fig:PCA:nocorr:snap:0.05} highlights some important points.
First, there is a clear distribution shift between the winter and summer months, in part due to the fact that electricity consumption is lower in summer than in the winter. Second, and most importantly, {\em the reconstructed distribution does not intersect the historical distribution.} Thus, unless an very large level of noise is used in the disaggregation, simply scaling the total load and applying noise will not capture the real distribution.
This is partly caused by the fact that the snapshot at hand is not representative of the considered historical distribution.
Indeed, Figure \ref{fig:PCA:nocorr:hist:0.05} displays the same disaggregation, this time using ratios estimated from the same month of the previous year. While the historical estimates yield a substantial improvement, the distributions still only partially overlap, especially for winter.

Overall, these observations indicate that generating realistic time series is not a trivial task: it requires a novel, principled approach to capture the realities of power system operations. 

\subsection{Contributions and Outline}

The goal of this paper is to generate synthetic times series data, in order to replicate the operations of a US TSO, e.g., MISO, on the French transmission system.
This requires times series for load and renewable production, as well as realistic generation offer bids that mimic the economics of a US transmission systems.

To address the above limitations, this paper proposes a data-driven methodology for reconstructing high-fidelity time series data that are spatio-temporally consistent. The proposed methodology has the following properties:
\begin{enumerate}
    \item it only requires a network snapshot and publically-available, aggregate TSO data;
    \item it reflects the variability of load, renewable output, and  production costs of conventional generators;
    \item it captures the spatio-temporal correlations between these time series;
    \item it is computationally efficient.
\end{enumerate}
\noindent

\noindent
The rest of the paper is organized as follows. Section \ref{sec:reconstruction} presents a principled methodology to reconstruct geo-spatial data from a network snapshot. Section \ref{sec:bids} shows how to map publicly-available generation offers and outage schedules to individual generators in the original system. Section \ref{sec:disaggregation} proposes a disaggregation procedure to recover time series at the individual component level while capturing spatial and temporal correlations. Section \ref{sec:extensions} discusses possible extensions and concludes the paper. Each step of the proposed methodology is demonstrated on the French transmission system, for which the approach generates synthetic, but highly realistic, time series at the component level. 

The paper uses the following notations. The set $\{1, 2, ..., N \}$ is denoted by $[N]$. Vectors are denoted in bold, e.g.,  $\mathbf{x}$.
The element-wise product between two vectors $\mathbf{x}, \mathbf{y}$ is denoted by $\mathbf{x} \odot \mathbf{y} = (x_1 y_1, x_2 y_2, ..., x_n y_n)$. The Kronecker product between two matrices $A, B$ is denoted by $A \otimes B$.

%% file: tex/reconstruction.tex
\section{Grid geodata reconstruction}
\label{sec:reconstruction}

Some test cases such as \cite{TAMUData} have been created to replicate existing grids, and therefore include realistic geodata, e.g., the location of every generator and/or substations.
However, most test cases available in \cite{Zimmerman2011_MATPOWER,coffrin2014nesta,babaeinejadsarookolaee2019power} do not.
Therefore, the first step in the proposed approach is to recover reasonable geo-coordinates for each component of the considered system.
This is critical to ensure that the subsequently-generated time series are spatially consistent.

The network data at hand consists of a snapshot of the French transmission grid provided by RTE, the French TSO.
It contains information about each bus, transmission line and transformer, generator and load. However, all component names are obfuscated, which prevents a direct identification. Geo-coordinates for each bus are then reconstructed by combining this network information with publicly-available data, namely, the location and voltage level of all substations in France \cite{ODRE}.
Note that, while similar information may not be available for all systems, other sources of information can be used in its place.
For instance, in the US, the Energy Information Administration (EIA) publishes information about all electricity-generating units \cite{EIA860}, including geo-coordinates.

The reconstruction of geo-coordinates is formulated as an optimization problem. Let $x_{i, j}$ be a binary variable that takes value $1$ if substation $i$ from the snapshot is mapped to location $j$, and $0$ otherwise. Then, for each snapshot substation $i$, let $S_{i}$ be the set of compatible geolocations, i.e., $j \in \S_{i}$ if and only if 1) geolocation $j$ is the same region as substation $i$, and 2) the voltage level at geolocation $j$ is compatible with that of substation $i$.
Also denote by $\tilde{D}_{i, i}$ the approximate length of transmission line $(i, i')$ in the snapshot; this quantity is evaluated by assuming that a line length is proportional to its resistance.
The reconstruction problem can be expressed as

\begin{subequations}
\label{eq:reconstruction}
\begin{align}
    \min_{\mathbf{x}} \quad &
    \sum_{i, i', j, j'} x_{i, j} x_{i', j'} \left(D_{j, j'} - \tilde{D}_{i, i'} \right)^{2} \\
    s.t. \quad
    \label{eq:reconstruction:compatible_location}
    & \sum_{j \in S_{i}} x_{i, j} = 1, \quad \forall i,\\
    \label{eq:reconstruction:no_dupplicates}
    & \sum_{i} x_{i, j} \leq 1, \quad \forall j,\\
    & x_{i, j} \in \{0, 1\},
\end{align}
\end{subequations}
where the objective seeks to minimize the reconstruction error on the length of transmission lines, constraint \eqref{eq:reconstruction:compatible_location} enforces that each substation is assigned to a compatible location, and constraint \eqref{eq:reconstruction:no_dupplicates} ensures that no two substations are assigned to the same location.

\begin{figure}[!t]
    \centering
    \includegraphics[width=0.90\columnwidth]{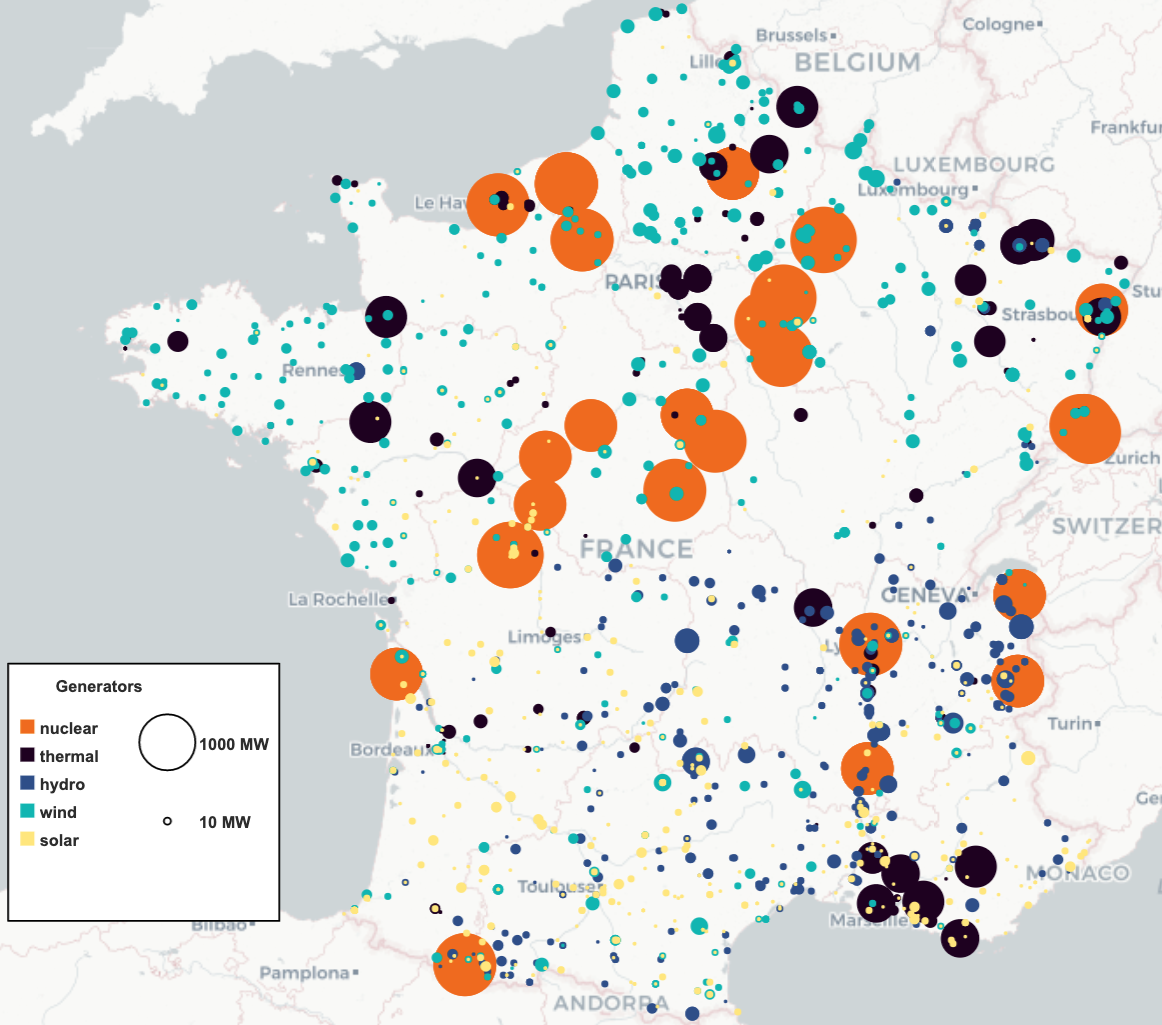}
    \caption{Reconstructed Locations of Generation Units in France.}
    \label{fig:geodata:France}
\end{figure}

Problem \eqref{eq:reconstruction} is a quadratic binary optimization problem.
While off-the-shelf solvers such as CPLEX and Gurobi can solve it exactly, doing so is not tractable due to the large number of variables and the presence of non-convexities.
In addition, Problem \eqref{eq:reconstruction} is only a proxy for reconstructing geocoordinates, i.e., an optimal solution is not needed in this case.
Therefore, a simple local search heuristic is implemented, which finds acceptable solutions in a few seconds of computing time.
The reconstructed locations for generators in France is shown in Figure \ref{fig:geodata:France}. While the reconstruction is approximate, the distribution of nuclear generators for instance is similar to that of the real nuclear power plants in France.

Note that the information needed to specify Problem \eqref{eq:reconstruction} is 1) line resistances, and 2) regional and voltage information for the buses.
This information is present in all publicly-available OPF test cases \cite{Zimmerman2011_MATPOWER,babaeinejadsarookolaee2019power,TAMUData}, which allows the proposed reconstruction to be conducted systematically.
The structure of Problem \eqref{eq:reconstruction} can easily be adapted to other settings, e.g., those where only locations of real generators are known, by mapping generators to locations, instead of substations.
Generator limits and fuel types can then be used to establish compatibility constraints between generators and geolocations.

%% file: tex/bids.tex
\section{Generation offer bids}
\label{sec:bids}

\begin{figure}[!t]
    \centering
    \includegraphics[width=0.9\columnwidth]{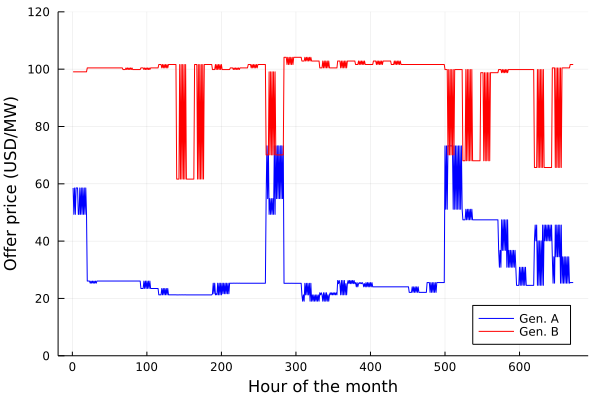}
    \caption{Energy Offer Price for Two Generators in PJM in February 2018.}
    \label{fig:generation_offer_price}
\end{figure}

Because TSOs operate power systems to minimize overall cost, variability in production costs affect commitment and dispatch decisions.
In systems like MISO and PJM, market participants submit hourly bids which include offer prices and economic limits, in addition to commitment-related data such as minimum up- and down-time, as well as startup delays and associated costs. Figure \ref{fig:generation_offer_price} displays the evolution of  energy offer prices for two generators over the month of February 2018.
Both prices fluctuate over the month, and sudden variations are not uncommon within one day. The magnitude of the price variations is also quite large: for generator A, the offer prices vary between 20 and almost 80 \$/MW, a four-fold increase. Therefore, in order to accurately replicate a TSO's operation over multiple days, this variability must be taken into account.

In the US, in order to ensure market transparency, TSOs publish anonymized generation and demand bids \cite{MISOData,PJMDataMiner}.
This data can then be mapped onto the generators present in the network snapshot, thereby recreating realistic bidding behavior across time.
To avoid confusion, in the rest of this section, ``generator" shall refer to a generator in the network snapshot at hand, while ``market participant" shall refer to an entity that bids into the considered market. Thus, the reconstruction consists in matching each generator with a market participant. First, the fuel type and maximum output of each market participant is identified. While the latter is simply estimated from the largest observed bid, the second can be approximately guessed by combining market offer data with public generator information, e.g., from \cite{EIA860}.
For instance, a market participant in PJM submitting who submits a bid for 500MW may only be mapped to an EIA generator in the PJM system, whose maximum output is at least 500MW.
Note that several systems in \cite{TAMUData} already include the EIA plant number of each generator, which simplifies the reconstruction.
Once each market participant is assigned a fuel type and maximum output, it can be mapped to a generator in the snapshot with same fuel type and similar maximum output.

The main limitation of this approach, is that the transmission system from which bids are collected may have a different fuel mix than the network snapshot at hand. For instance, in the US, most generators use coal or natural gas, while in France, nuclear generation accounts for most of the installed capacity. To alleviate this effect, the same market participant may be mapped to multiple generators.

%% file: tex/disaggregation.tex
    \section{Time series disaggregation}
\label{sec:disaggregation}

Realistic time series for individual loads and renewable generators are obtained via a disaggregation procedure.
Publicly-available historical load and electricity production are collected at a system-wide level with sub-hourly granularity.
These system-wide time series are then disaggregated into individual times series, taking into account spatio-temporal correlations between individual components of the network, and respecting (ground truth) regional totals.

The French Transmission Operator (RTE) provides publicly  real-time information on the total output of each type of generator type (e.g solar,  wind among others) and power consumption \cite{eCO2mix}. The system is partitioned in 12 regions and the data is provided at a regional level. In particular, for each of the 12 regions and at a 30-minute granularity, information on (solar and wind) production and consumption may be collected over a whole year. The main assumption underlying the disaggregation  is that, within a small geographical region, individual solar, wind, and load behavior follow the same trend with some individual spatio-temporal volatilities. 

\begin{figure}[!t]
    \centering
    \includegraphics[width=\columnwidth]{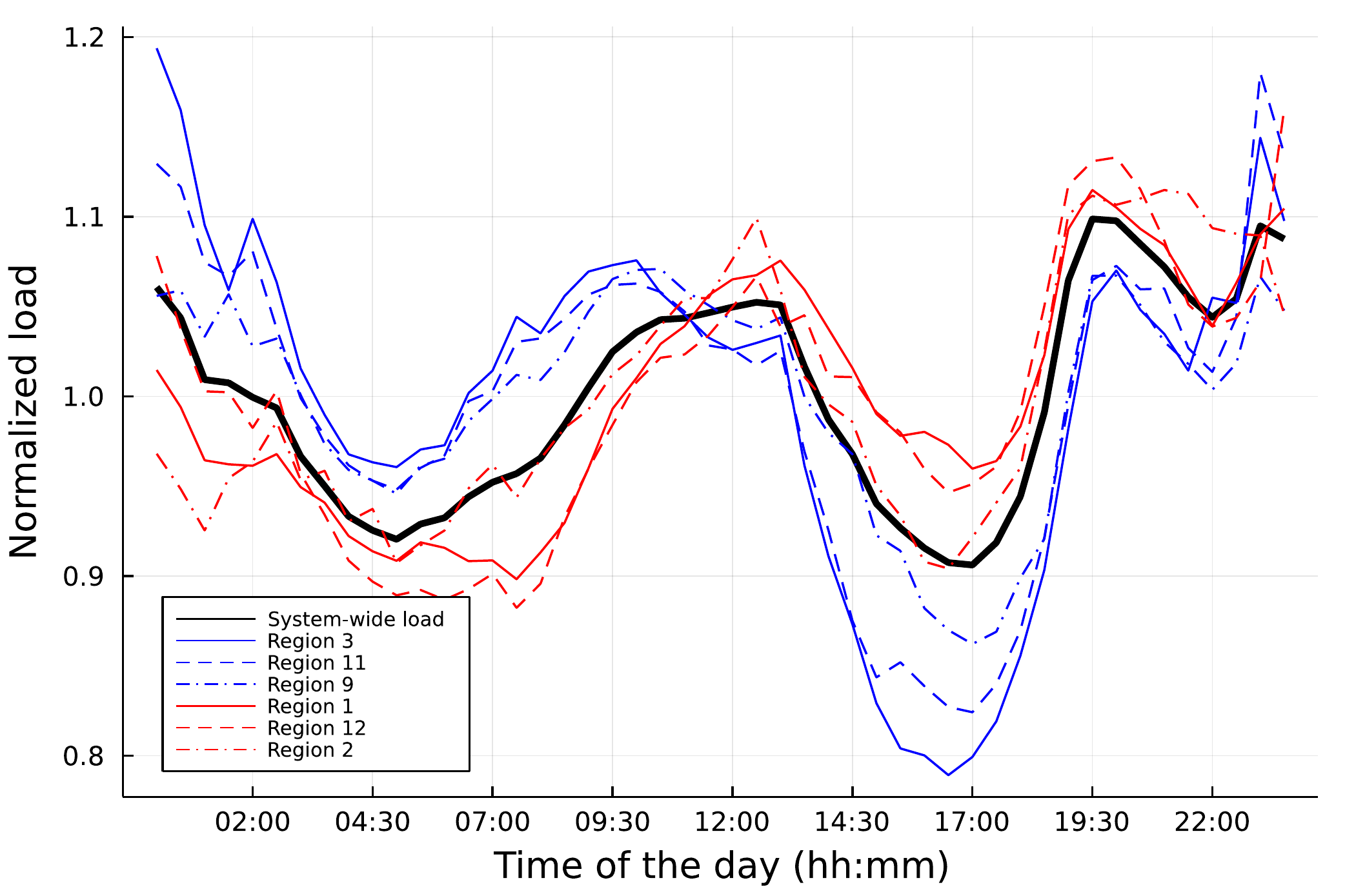}
    \caption{Historical System-wide Level Load Compared to Regional Loads.}
    \label{fig:regional_loads}
\end{figure}

It is possible to verify an analogous assumption between the system-wide load and the regional loads. Figure \ref{fig:regional_loads} compares the historical system-wide load to the historical load of six French regions on a single day. Regions 3, 11, and 9 (resp. 1, 12, and 2) are located in the West (resp. East) part of mainland France. The following three observations are in order:
\begin{enumerate}
    \item The regional loads follow the trend of the system-wide load with additional volatilities.
    \item The regions in the West (resp. East) behave very similarly. Thus the volatilities present spatial correlations.
    \item Regional loads keep a higher or lower normalized value compared to the system-wide load for several hours. Hence, the volatilities present temporal correlations.
\end{enumerate}

\subsection{Generating Spatio-Temporal Volatilities}

This section presents the disaggregation procedure for power consumption: the same procedure is applied to solar and wind. The procedure could also be applied by considering a unified vector of solar, wind and load, but is unnecessary since the correlations between solar, wind and load are already captured by the regional ground truth values. The following terms are defined:
\begin{itemize}
    \item $\mathcal{T}$: The set of time periods. $T = |\mathcal{T}| $.
    \item $\mathcal{R}$: The set of regions. $R = |\mathcal{R}|$.
    \item $\mathcal{N}_r$: The set of individual loads at region $r \in \mathcal{R}$.
    \item $\mathcal{N} = \cup_{r \in \mathcal{R}} \mathcal{N}_r$, the set of loads. $N = |\mathcal{N}|$
    \item $L_{r}^t$: Load realization for region $r \in \mathcal{R}$ at time $t \in \mathcal{T}$.

\end{itemize}
Moreover, the percentage that each individual load contributes to the regional load is defined as:
\begin{align}
\boldsymbol{p}_r = [p_{r,1}, p_{r,2}, ..., p_{r, N_r}] & \quad r \in \mathcal{R}
\end{align}
These vectors can be estimated directly from a single topology snapshot where a single load profile is available.
Based on this, the following disaggregation procedure is proposed:
\begin{align}
    \hat{\boldsymbol{L}}_r^t = \frac{L_r^t}{(\boldsymbol{p}_r)^{\top} \boldsymbol{y}_r^t} (\boldsymbol{p}_r \odot \boldsymbol{y}_r^t )& \quad r \in \mathcal{R}, t \in \mathcal{T} \label{good_load}
\end{align}
where $\boldsymbol{y}^t = [\boldsymbol{y}_1^t, \boldsymbol{y}_2^t, ..., \boldsymbol{y}_R^t]$ is a random vector realization that captures the aforementioned volatility for a single time $t$. The normalization term $(\boldsymbol{p}_r)^{\top} \boldsymbol{y}_r^t$ ensures that the total load in the region is equal to the ground truth. A number of properties are desirable for the random variable $ \tilde{{\boldsymbol y}}_r^t = [\tilde{y}_{r,1}^t, \tilde{y}_{r,1}^t, ..., \tilde{y}_{r,N_r}^t]$:
\begin{enumerate}
    \item $\mathbb{E}[ \tilde{{\boldsymbol y}}_r^t]$ = 1, i.e., on expectation, the individual loads follow the regional trend.
    \item $ \tilde{{\boldsymbol y}}_r^t \geq 0$. 
    \item The distribution should be unimodal around the mean. This makes it unlikely that individual load values will diverge significantly from the regional load, but extreme cases should still appear in the dataset. 
    \item $\tilde{{\boldsymbol y}}_{r,i}^t$ should capture spatial correlations between individual loads. For instance, residential and commercial loads located geographically nearby should demonstrate similar behavior (e.g., due to the spatial correlation of temperature and consistency of people's activities). Also $\tilde{{y}}_{r,i}^t, \tilde{{ y}}_{r,i}^{t+1}$ should be temporally correlated to make extreme changes in a short time interval unlikely.
\end{enumerate}
The Log-normal distribution with the appropriate parameters can be used to generate coefficients that satisfy properties 1), 2) and 3). Log-normal distribution was also used in \cite{donno2019} to generate load profiles for the French power grid, but without capturing spatial and temporal correlations. 

The coefficients $\boldsymbol{y}_r^t$ can be generated in order to satisfy property 4) as well by leveraging the geographical information on the topology. The pair-wise distances between the locations of all individual loads in the system are calculated using the geocoordinate information from Section II. This matrix is denoted as $D$ with $D_{ij}$ being the distance between loads $i, j$. The spatial covariance matrix $\Sigma_1$ is defined as:
\begin{align}
    (\Sigma_1)_{i,j} = \alpha \exp(-\frac{D_{ij}^2}{2 \sigma^2}) \quad \forall i,j \in \mathcal{N} \times \mathcal{N}
\end{align}
$\Sigma_1$ is known as a Radial basis function kernel and is widely used to capture correlations based on distances between elements (see \cite{3569}, Chapter 4). A small value for $D_{i,j}$ leads to a high correlation between elements $i$ and $0j$. The term $\alpha > 0$ controls the variance of the components while $\sigma$ controls the sensitivity of the correlations based on the distances. Higher values of $\sigma$ lead to stronger spatial correlations. Similarly, the temporal correlation matrix $\Sigma_2$ is defined as:
\begin{align}
    (\Sigma_2)_{i,j} = \exp(- \theta |t_i - t_j|) \quad \forall i,j \in \mathcal{T} \times \mathcal{T}.
\end{align}
The spatio-temporal covariance matrix $\Sigma$ can be specified by the Kronecker product of these two matrices, i.e.,
\begin{align}
    \Sigma = \Sigma_1 \otimes \Sigma_2
\end{align}
that has size $N T \times N T$.

To compute the load disaggregation through Equation (\ref{good_load}), it remains to generate a matrix $Y = [\boldsymbol{y}^1, \boldsymbol{y}^2, ..., \boldsymbol{y}^T]$, sized $N \times T$, of coefficients. This can be achieved by sampling the distribution $\mathcal{N}(0, \Sigma_1 \otimes \Sigma_2)$ to obtain a vector $Y_0$ of size $NT$, which can then be converted to a multivariate Log-Normal distribution with the proper mean using 
\[
Y_i = \exp((Y_0)_i) + 1 - \exp(\frac{1}{2}\Sigma_{ii}), \forall i \in [NT].
\]
To generate these coefficients efficiently, and avoid computing the gigantic Kronecker product, the disaggregation method follows the following steps. First, it generates a matrix $X_{N \times T} \sim$ $\mathcal{N}(0, I_N \otimes I_T)$ of uncorrelated $\mathcal{N}(0,1)$ samples using
the matrix Normal Distribution \cite{gupta1999matrix}. Second, it computes Cholesky Decompositions of $\Sigma_1 = AA^{\top}$ and $\Sigma_2 = B^{\top}B$ of $\Sigma_1$ and $\Sigma_2$.
It follows that the matrix $Y_0$ can be computed as $A X B$ and has distribution $\mathcal{N}(0, \Sigma_1 \otimes \Sigma_2)$. The random vector $Y$ defined previously follows a multivariate Log-Normal distribution with:
\begin{flalign*}
  &  \mu_i = 1, \forall i \in [NT]\\
  & \Sigma'_{ij} = \exp(\frac{1}{2}(\Sigma_{ii} + \Sigma_{jj}))(\exp(\Sigma_{ij})-1) 
\end{flalign*}

\begin{figure}[!t]
    \centering
    \includegraphics[width=\columnwidth]{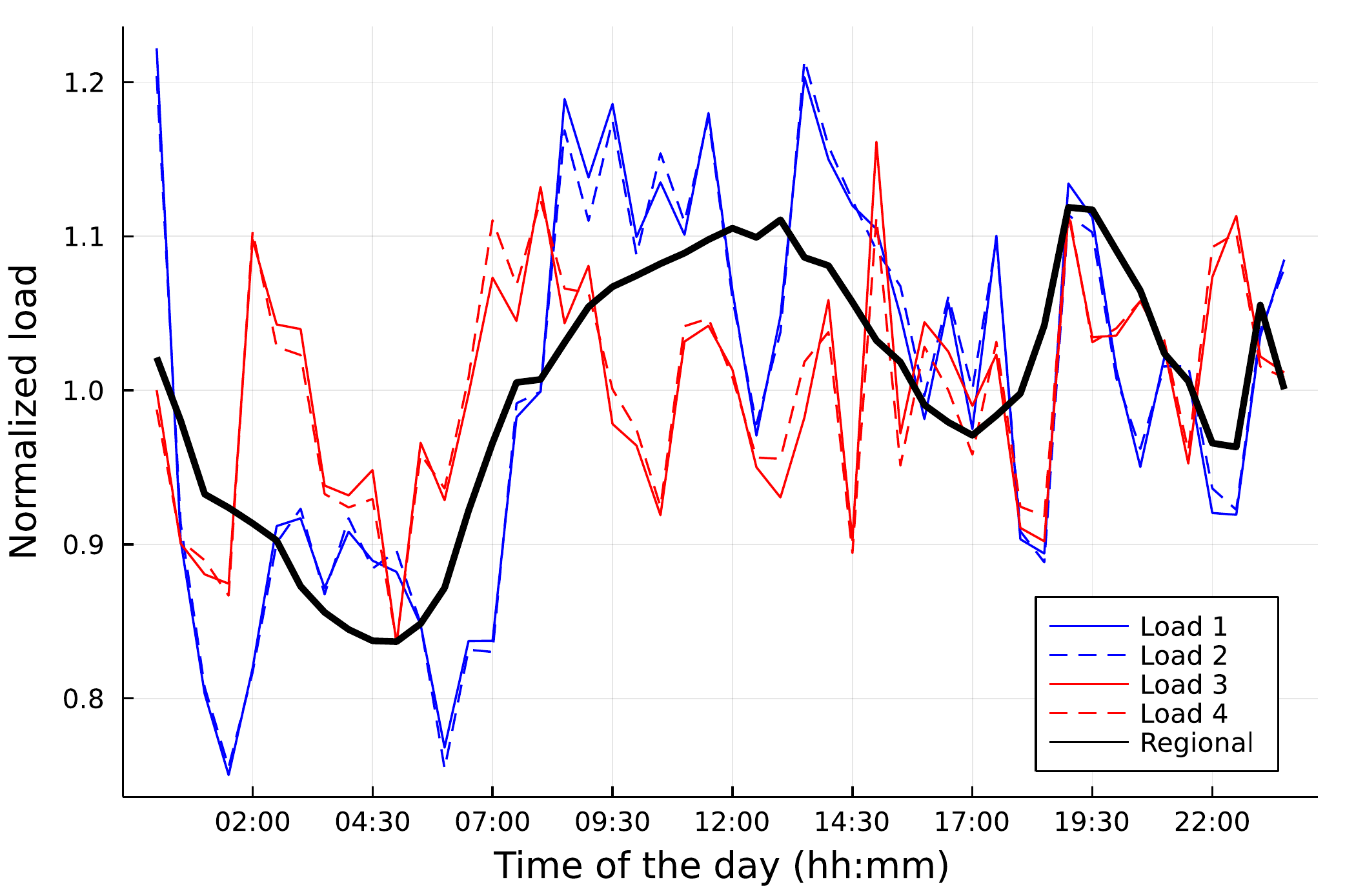}
    \caption{Regional and Disaggregated Load Profiles for the Occitanie Region.}
    \label{fig:bunchloads}
\end{figure}

The coefficients $\sigma, \theta, \alpha$ were set to $1,1,0.01$, respectively. The standard deviation of the individual coefficients is $$\sigma[Y_{i,j}] = \sqrt{\exp{(\frac{1}{2} 2 \alpha)}(\exp{(\alpha)} - 1)} \approx 0.1$$ A data-driven estimation of these parameters would require data at a finer granularity, which a TSO could provide. Moreover, different values of $\sigma, \theta, \alpha$ may be used for the solar and the wind disaggregation (e.g., wind output is more volatile than solar output, thus a higher value of $\alpha$ can be used for wind). If the granularity needed for time-series is smaller than the granularity of publicly available values, for instance every 5 minutes, a standard linear interpolation between subsequent time periods can be applied.

The result of the disaggregation procedure is illustrated in Figure \ref{fig:bunchloads} for a given region on a particular day. Loads 1, 2, 3, and 4 are located in the same region and follow the behavior of the regional load. Loads 1 and 2 are  located nearby geographically and are far apart from loads 3 and 4 which are also close to each other. Loads 1 and 2 (resp. 3 and 4) behave similarly due to enforcement of spatial correlations.

\subsection{DC Feasibility Recovery}

Although the resulting time series display desirable statistical properties, preliminary experiments showed that they may yield test cases for which the DC-OPF is infeasible.
This is mostly unrealistic -- there has been no blackout in France in the recent years, and of limited practical interest, as it would automatically render a unit commitment or economic dispatch problem infeasible.
A deeper analysis revealed that such infeasible instances are typically caused by a few individual loads, located at the edge of the grid, whose increases cause the neighboring line(s) to become congested.

To overcome this limitation, a DC-feasibility recovery step is proposed, wherein individual loads are adjusted so as to ensure a feasible DC-OPF, while minimizing the deviation from the original disaggregated time series.
The latter ensures that spatio-temporal correlations are preserved.
This procedure is computationally equivalent to solving a DC-OPF, for every time period, and is therefore efficient.

\begin{figure}[!t]
    \centering
    \begin{subequations}
\begin{align}
    \min_{\bar{\boldsymbol{L}}^t} & \quad \lvert \lvert \bar{\boldsymbol{L}}^t - \hat{\boldsymbol{L}}^t \rvert \rvert_1 + \sum_{i=1}^{N} s_i^2 \label{lp-dc-obj} & \\
    \text{s.t} & \quad \bar{\boldsymbol{L}}^t \text{ is DC-feasible} \label{lp-dc-feas} &\\
    & \quad \sum_{i \in \mathcal{L}^r} (\bar{\boldsymbol{L}}^t)_i = \boldsymbol{L}_r^t & \forall r \in \mathcal{R} \label{lp-reg}
    \\ 
    & \quad (\bar{\boldsymbol{L}}^t)_i \leq \max((\hat{\boldsymbol{L}}^t)_i, (\boldsymbol{L}^0)_i) + s_i & \forall i \in \mathcal{N} \label{lp-1} \\
    & \quad (\bar{\boldsymbol{L}}^t)_i \geq \min((\hat{\boldsymbol{L}}^t)_i, (\boldsymbol{L}^0)_i) - s_i & \forall i \in \mathcal{N} \label{lp-2}
\end{align}
\end{subequations}
    \label{fig:dc_proj}
    \caption{The Optimization Model for DC Feasibility Restoration.}
\end{figure}

The optimization problem that models the feasibility recovery is displayed in Figure \ref{fig:dc_proj} for a given time $t \in \mathcal{T}$.
The input to the model is the disaggregated load profile $\hat{\boldsymbol{L}}^t$, the ground truth regional loads $\boldsymbol{L}^t_r$ and a single historical load value $\boldsymbol{L}^0$ (which is typically included in the network snapshot).
The problem aims at finding a DC-feasible load that is close to the disaggregated load $\hat{\boldsymbol{L}}^t$. Constraint (\ref{lp-dc-feas}) ensures that the resulting load $\boldsymbol{\bar{L}}^t$ is DC-feasible for the given network (i.e., there exists a dispatch within the given operating bounds that satisfies the load).
These constraints are the standard DC-OPF constraints which include Nodal Power Balance, and Branch Thermal limits.
Moreover, in order to avoid artificial congestions, the thermal limits are reduced by a small factor (e.g., 5\%).
When this reduction was not applied, several unrealistic cases were noticed where lines are congested only due the existence of a load that matches the thermal capacity of the adjacent line.
Constraint (\ref{lp-reg}) ensures that the resulting regional loads will match the ground truth given by the publicly available data.
Finally, Constraints (\ref{lp-1}), (\ref{lp-2}) penalize extreme variations outside of the interval defined by the nominal load and the disaggregated load. 

\begin{figure}[!t]
    \centering
    \includegraphics[width=0.90\columnwidth]{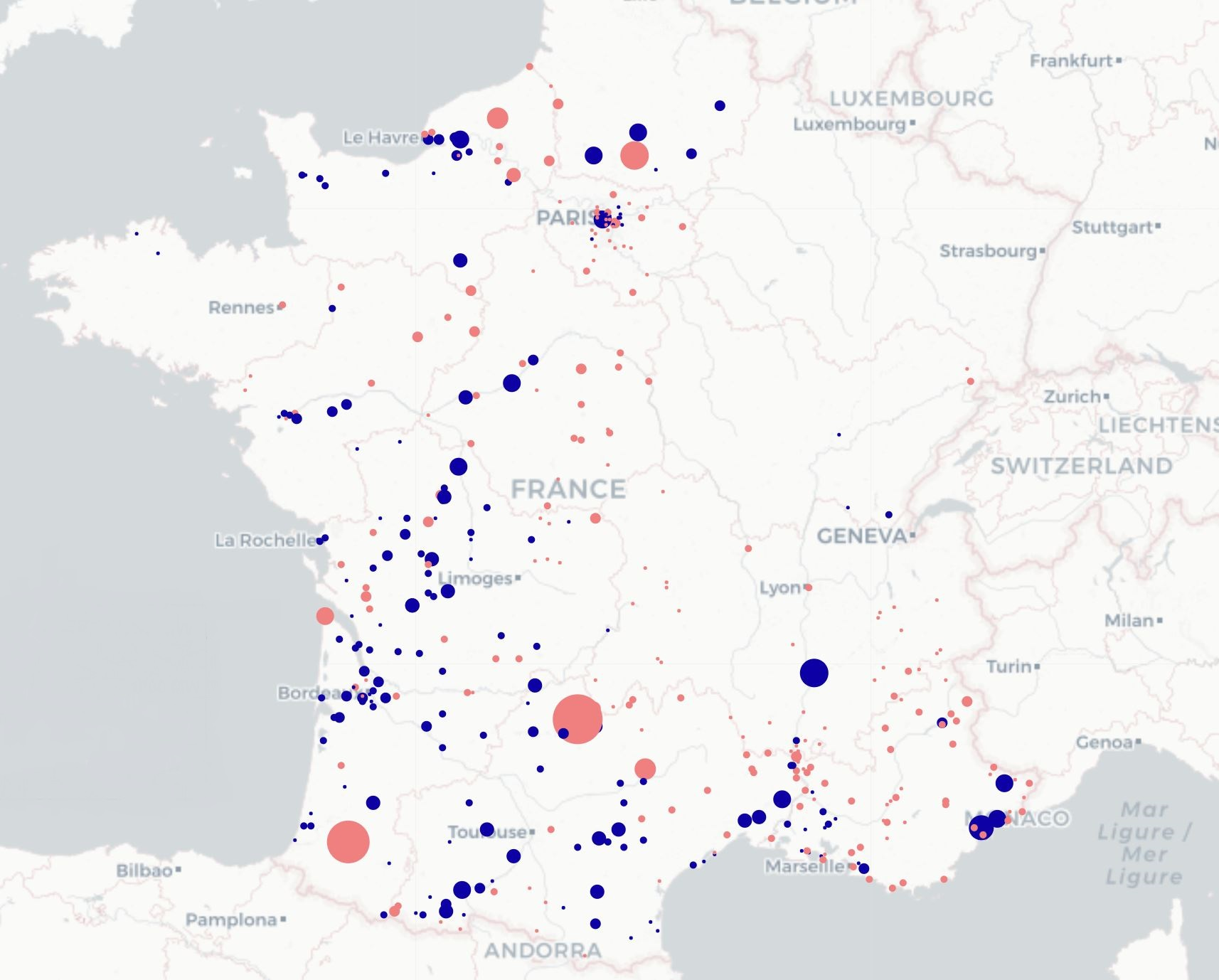}
    \caption{The difference between the DC-feasible load and the disaggregated load at 9pm on February 28, 2018. Red (resp. blue) indicates an algebraic increase (res. decrease); the area of each circle is proportional to the magnitude of the difference.}
    \label{fig:diff_loads}
\end{figure}

An example of the difference between the disaggregated load $\hat{\boldsymbol{L}}_t$ and the result of the optimization $\bar{\boldsymbol{L}}_t$ can be viewed in Figure \ref{fig:diff_loads} for a given time on February 28, 2018, which is one of the highest load days in France in 2018 with the peak being over $95000$ MW.
The figure displays the loads that are decreased (blue) or increased (red) due to the feasibility restoration.
The total difference between the two loads due to the feasibility restoration is $7007$ MW indicating the necessity of the procedure for providing a realistic load profile.
The feasibility restoration significantly alters the value on a small number of loads (big dots) in order to ensure DC-feasibility, with the largest individual difference to be almost 1000 MW. 
To ensure that the regional loads remain the same, several loads in the regions are changed appropriately.
In general, it was observed that the largest changes due to the feasibility restoration are on high load days (Winter) while on low load days (Summer) the changes are minimal.

\subsection{Computational and Accuracy Results}

Time series for the solar and wind outputs, and loads in the French Transmission system are generated independently for the years 2017-2020, with 5-minute granularity.
The total number of individual renewable generators and loads in the network snapshot at hand is around 8000.
The generation of the spatio-temporal volatilities and the resulting disaggregated time-series takes less than 30 minutes on a standard laptop.
Moreover, each execution of the DC feasibility restoration takes around 2 seconds and $24 \times 12 = 288$ executions are needed per day.
The executions per day were run in parallel using the PACE High-Performance Cluster at the Georgia Institute of Technology.

\begin{figure}[!t]
    \centering
    \includegraphics[width=0.95\columnwidth]{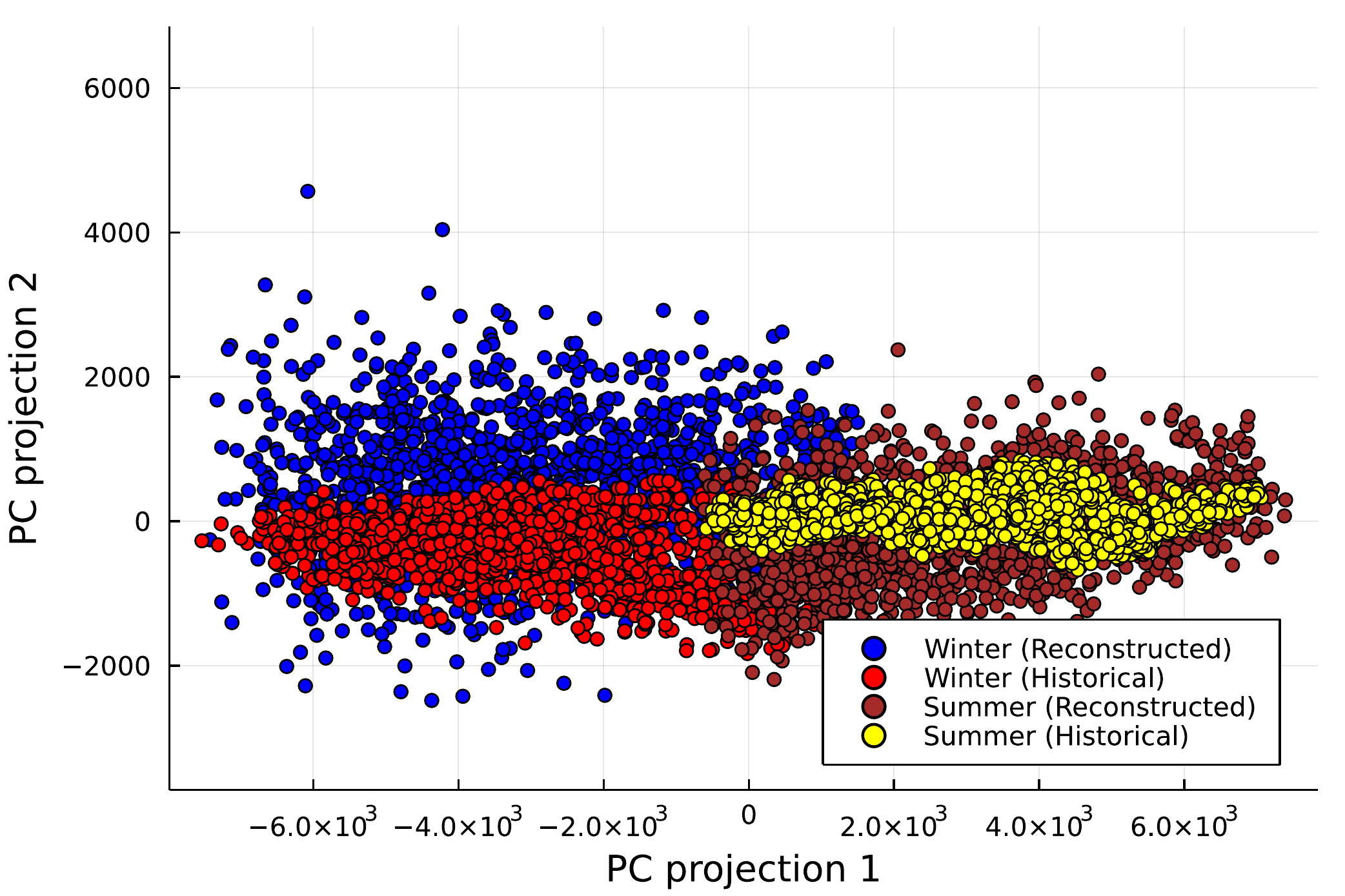}
    \caption{Projected Distribution of Regional Load Profiles Obtained from Historical Data, and from a Uniform Disaggregation as in \cite{donno2019}, using scaling ratios  estimated from the previous year's data and $10\%$ noise.
    The two principal components used for the projection capture $90.7\%$ of the total variance.}
    \label{fig:PCA:nocorr:hist:0.1}
\end{figure}

\begin{figure}[!t]
    \centering
    \includegraphics[width=0.95\columnwidth]{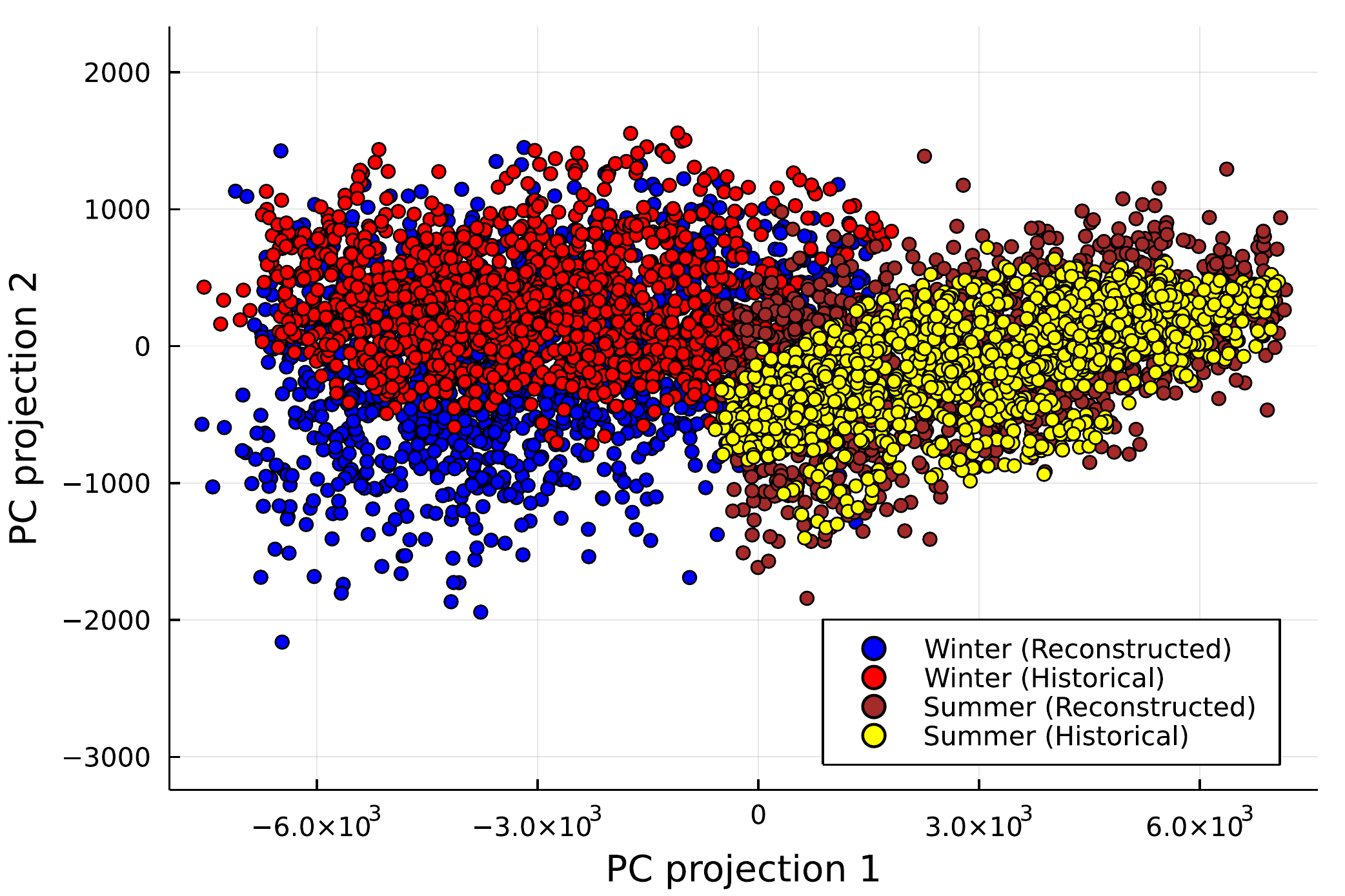}
    \caption{Projected Distribution of Regional Load Profiles Obtained from Historical Data, and from the proposed method, using scaling ratios estimated from the previous year's data and $10\%$ noise.
    The two principal components used for the projection capture $96.5\%$ of the total variance.}
    \label{fig:PCA:corr:hist:0.1}
\end{figure}

\begin{figure}[!t]
    \centering
    \includegraphics[width=0.95\columnwidth]{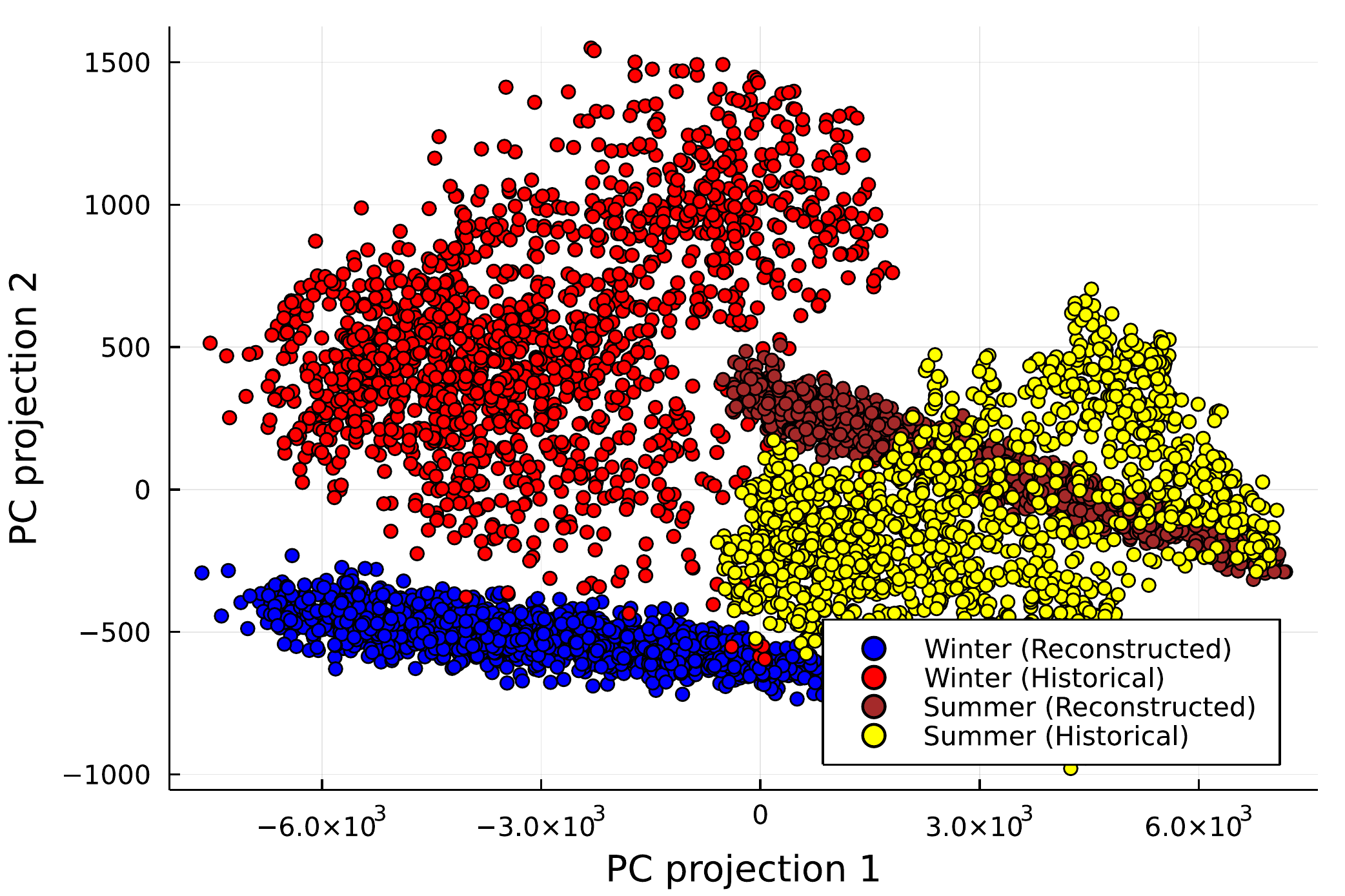}
    \caption{Projected Distribution of Regional Load Profiles Obtained from Historical Data, and from the proposed method, using scaling ratios estimated from the previous year's data and $5\%$ noise.
    The two principal components used for the projection capture $97.5\%$ of the total variance.}
    \label{fig:PCA:corr:hist:0.05}
\end{figure}

It is also important to demonstrate the benefits of introducing correlations in the reconstruction. Since the above data generation follows the historical regional time series by design, Figures \ref{fig:PCA:nocorr:hist:0.1}-\ref{fig:PCA:corr:hist:0.05} illustrate the methodology on the disaggregation of national into regional load profiles. On the one hand, Figure \ref{fig:PCA:nocorr:hist:0.1} uses the same uniform disaggregation as in Figure \ref{fig:PCA:nocorr:hist:0.05}, this time with $10\%$ noise to better overlap the real distribution.
On the other hand, Figures \ref{fig:PCA:corr:hist:0.1} and \ref{fig:PCA:corr:hist:0.05} use the proposed, correlation-based disaggregation with $10\%$ and $5\%$ noise, respectively. While the uniform scaling-based reconstruction does overlap with the real distribution, a significant proportion falls outside it.
In contrast, when taking correlations into account, the reconstruction with $10\%$ noise (Figure \ref{fig:PCA:corr:hist:0.1}) overlaps almost perfectly with the real distribution in the summer, and the number of outliers in the winter is significantly reduced compared to the uncorrelated disaggregation. Finally, Figure \ref{fig:PCA:corr:hist:0.05} corroborates the earlier findings of Figure \ref{fig:PCA:nocorr:snap:0.05}: unless a sufficient amount of noise is introduced, the synthetic data will remain concentrated around whichever reference point is used, and likely fail to capture the real distribution.

%% file: tex/extensions.tex
\section{Conclusion}
\label{sec:extensions}

The paper has presented quantitative evidence that spatio-temporal correlations should be taken into account when generating synthetic time-series data.
Simple data augmentation strategies should take advantage of regional time series, whenever they are available.
A principled methodology has been proposed, which only requires publicly-available data, and is able to generate time series data that are spatio-temporally consistent.
The approach has been illustrated on the French transmission system, for which several years of realistic, synthetic time series data have been reconstructed.
Because it is computationally efficient, it can also be used to generate training data in the context of ML studies, or Monte-Carlo scenarios for stochastic optimization.

Given the appropriate historical data, several extensions are possible.
First, it is a well-known fact that residential and industrial loads behave differently.
Capturing this would require identifying which load buses belong to which class, which can only be done approximately when handed a single network snapshot with no geocoordinates.
However, such approaches are highly relevant when building synthetic grids, as described in \cite{Li2018_LoadModelingSynthetic,Li2021_CreationValidationLoadTimeSeries}. Similarly, wind and solar production are best estimated by taking into account weather information.
While there exist public databases of solar irradiance and (reconstructed) wind measurement \cite{Gilman2015_sam,DRAXL2015_WindIntegrtionToolkit}, these may not cover the geographic footprint of temporal period of interest. Naturally, in that context, accurate geodata reconstruction is critical.

\section*{Acknowledgments}
This research is partly funded by NSF Award 1912244 and ARPA-E Perform Award AR0001136.